\documentclass[conference]{IEEEtran}
\IEEEoverridecommandlockouts

\makeatletter
\let\old@ps@headings\ps@headings
\let\old@ps@IEEEtitlepagestyle\ps@IEEEtitlepagestyle
\def\confheader#1{%
	\def\ps@headings{%
		\old@ps@headings%
		\def\@oddhead{\strut\hfill#1\hfill\strut}%
		\def\@evenhead{\strut\hfill#1\hfill\strut}%
	}%
	\def\ps@IEEEtitlepagestyle{%
		\old@ps@IEEEtitlepagestyle%
		\def\@oddhead{\strut\hfill#1\hfill\strut}%
		\def\@evenhead{\strut\hfill#1\hfill\strut}%
	}%
	\ps@headings%
}
\makeatother

\confheader{%
	2020 28\textsuperscript{th} Iranian  Conference on Electrical Engineering (ICEE)
}
\usepackage[pscoord]{eso-pic}
\newcommand{\placetextbox}[3]{
	\setbox0=\hbox{#3}
	\AddToShipoutPictureFG*{ \put(\LenToUnit{#1\paperwidth},\LenToUnit{#2\paperheight}){\vtop{{\null}\makebox[0pt][c]{#3}}}
	}
}
\placetextbox{.23}{0.055}{\small{978-1-7281-7296-5/20/\$31.00~\copyright 2020 IEEE}}

\usepackage{cite}
\usepackage{amsmath,amssymb,amsfonts}
\usepackage{algorithmic}
\usepackage{graphicx}
\usepackage{textcomp}
\usepackage{amsmath}
\usepackage{xcolor}
\usepackage{url}
\usepackage{multirow}
\usepackage{subcaption}
\usepackage[numbers]{natbib}
\begin{document}

\title{A Machine Learning Approach to Persian Text Readability Assessment Using a Crowdsourced Dataset}

\author{\IEEEauthorblockN{Hamid Mohammadi}
\IEEEauthorblockA{\textit{Computer Engineering Department} \\
\textit{K. N. Toosi University of Technology}\\
Tehran, Iran \\
mohammadi2823@email.kntu.ac.ir}
\and
\IEEEauthorblockN{Seyed Hossein Khasteh}
\IEEEauthorblockA{\textit{Computer Engineering Department} \\
\textit{K. N. Toosi University of Technology}\\
Tehran, Iran \\
khasteh@kntu.ac.ir}
}

\maketitle

\begin{abstract}
An automated approach to text readability assessment is essential to a language and can be a powerful tool for improving the understandability of texts written and published in that language. However, the Persian language, which is spoken by over 110 million speakers\footnote{“Persian language,” Sep 2018. [Online]. Available:https://en.wikipedia.org/wiki/Persian\_language}, lacks such a system. Unlike other languages such as English, French, and Chinese, minimal research studies have been conducted to develop an accurate and reliable text readability assessment system for the Persian language.

In the present research, the first Persian dataset for text readability assessment was gathered, and the first model for Persian text readability assessment using machine learning was introduced. The experiments revealed that this model was accurate and could assess the readability of Persian texts with a high degree of confidence. The results of this study can be used in several applications such as medical and educational text readability evaluation and have the potential to be the cornerstone of future studies in Persian text readability assessment.
\end{abstract}


\section{Introduction}
\label{Introduction}
With the advent of the World Wide Web, the volume of digital content such as text is growing fast every day. One of the main properties of a text is its readability. Readability or difficulty of a text signifies how understandable it is for a human reader. The massive amount of accessible texts makes it hard to find those texts with a certain readability level or accurately assess the readability of available texts by a human. With that in mind, the Internet is not the only place where we need to assess the readability of a text. Readability measurement is indispensable in an education system where it can help teachers/instructors find suitable content for students to read based on their reading skills or help textbook authors to evaluate their books in terms of suitability for the intended students. Second language learners can also benefit from an automated text readability assessment system to find suitable texts for their educational purposes \cite{xia2019text}. Another application of readability measurement lies in medical texts. Studies have shown that the readability of a text can significantly enhance its understandability by readers \cite{leroy2013user}. By measuring the readability of educational, medical contents written for patients, it is possible to ensure that it is understandable for the public. Text readability measurement has many other applications in areas such as advertising \cite{pancer2019readability}, publishing, and other related practices. Computers can help us facilitate the process of text readability assessment; however, we need an accurate and reliable measure to assess text readability.

Early automated text readability assessments were undertaken using readability formulae. These formulae would measure the readability of a text-based on some simple characteristics. One of the most popular and widely used readability formulae is the Flesch-Kincaid readability formula \cite{kincaid1975derivation}. This formula measures the readability of a given text regarding the total number of words and syllables in sentences. Nonetheless, readability formulae have been considered not accurate enough through several research studies \cite{heydari2012readability, begeny2014can, crossley2017predicting}. A more effective approach to the assessment of the text readability is to use machine learning techniques for this task. Therefore, in recent years, several research studies have been conducted on designing and testing a text readability assessment system using machine learning techniques. One of the first studies of this kind was the work of \citet{schwarm2005reading}, which introduced a model for English text readability assessment using machine learning approaches.

The only known method for Persian text readability assessment is the Flesch-Dayani \cite{dayani2000criteria} formula. Flesch-Dayani formula is a recalculated version of the Flesch-Kincaid formula, which makes it optimal for the Persian language.

As mentioned earlier, the traditional approaches to text readability assessment have not been accurate enough. However, no previous machine learning model is available for Persian language text readability assessment. The main reasons for the absence of such a model are the lack of any text readability datasets for the Persian language. In this research paper, a text readability dataset for the Persian language is gathered using a novel method. Further, the first machine learning model for Persian text readability assessment is designed and tested.

The structure of this paper is designed as follows: (i) We discuss the previous research studies performed in this field (section \ref{Related Works}); (ii) Our approach to gathering a Persian text readability dataset and its characteristics are proposed (section \ref{Dataset}); (iii) the details of a machine learning model for Persian text readability assessment is explained (section \ref{Proposed Approach}); (iv) Test results are presented (section \ref{Experiments}); and (v) The conclusions and future directions are discussed (section \ref{Conclusions and Future works}).

\section{State of the Art}
\label{Related Works}
The previously published research studies in the field of text readability assessment for the Persian language are limited. As mentioned before, the only well-known text readability measure the Persian language is Flesch-Dayani \cite{dayani2000criteria} measure. This formula is shown in Eq. \ref{eq:dayani}.

\begin{multline}
Flesch\text{-}Dayani\ Score = \\ 262.835 -0.846\cdot\frac{|letters|}{|words|}\ -\ 1.01\cdot\frac{|words|}{|sentences|}\
\label{eq:dayani}
\end{multline}

On the contrary, there are many research studies conducted in the field of text readability assessment for other languages such as English, French, and Chinese.

Approaches to text readability assessment can be classified into two major classes: (i) traditional approaches to text readability assessment and (ii) machine learning approaches. In this respect, traditional approaches are composed of simple variables and metrics and are easy to compute. In other words, these approaches mostly use surface features of a text to assess its readability. In contrast, machine learning approaches employ complex and deep features alongside traditional surface features to extract more information from the text.

One of the earliest research studies of traditional text readability assessment was the work of \citet{flesch1943marks}. Flesch published a readability measurement formula which was developed under contract with the U.S. Navy \cite{kincaid1975derivation}. Later it was recalculated under the name of Flesch-Kincaid formula and became one of the most popular text readability measures. Currently, many popular text processing programs are using this formula as a built-in readability assessment criterion\footnote{"Flesch–kincaid readability tests," Sep 2018. [Online]. Available: \url{https://en.wikipedia.org/wiki/Flesch-Kincaid_readability_tests}}. This formula is presented in Eq. \ref{eq:flesch}.

\begin{multline}
Flesch\text{-}Kincaid\ Grade Level = \\ 0.39 \cdot\frac{|words|}{|sentences|}\ +\ 11.8 \cdot\frac{|syllables|}{|words|}\ -\ 15.59
\label{eq:flesch}
\end{multline}

As shown in Eq. \ref{eq:flesch}, the Flesch-Kincaid formula only considers the number of syllables, words, and sentences for readability assessment. Lexile \cite{stenner1996measuring} is another text readability measure. It was the first readability measure that applied word frequency as a feature to measure the readability of a text.
Another research study that employed the statistical properties of a text to predict the readability level was the work of \citet{collins2005predicting}. Collins used a unigram language model to measure the text readability.

Despite the simplicity of traditional approaches, they lack the accuracy required to assess the readability of a text reliably. Indeed, the main problem of traditional approaches is that they only take into account a minimal number of features of a text to assess the readability \cite{petersen2009machine,hartley2016time}. The use of a limited set of features makes these algorithms less accurate. The other weakness of traditional approaches is that they need a long text to reach an accurate conclusion about the readability of a text \cite{kidwell2011statistical}. This problem, for example, can make such approaches unable to reliably assess the readability level of texts in snippets, short chats, or other applications of short texts. A low degree of accuracy of these approaches in assessing the readability of web pages is yet another drawback of these approaches \cite{collins2005predicting, petersen2006assessing, feng2009cognitively}. Generally, there are a significant number of research studies which have compared the human judgment with readability formulae. They have concluded that traditional assessment of text readability can have a vast difference from human\textquotesingle s evaluation \cite{heydari2012readability, begeny2014can, crossley2017predicting}.

The next class of approaches towards text readability assessment includes machine learning-based approaches. They outperform traditional approaches thanks to their intensive use of natural language processing (NLP) features and machine learning techniques \cite{franccois2012nlp}. Text readability assessment can be either regression or a classification problem; however, research studies have suggested that classification approaches can result in a better assessment of text readability \cite{feng2010comparison}.

One of the most common classifiers used in text readability assessment studies is Support Vector Machine (hereafter SVM). The reason behind this choice is that in order to assess the readability of a text accurately, many features should be extracted from the text, which will increase the dimensionality of the classifier\textquotesingle s input. On top of that, SVM naturally performs better on data with a high level of dimensionality compared to other classifiers such as neural networks. With that in mind, SVM seems to be an appropriate choice for text classification.
The most noticeable difference between the research studies on text readability assessment using machine learning techniques is the set of features each study has utilized. Selecting features depends on the number of criteria such as the text\textquotesingle s application, language, and many other parameters.

One of the first attempts to apply machine learning for text readability assessment was the works by \citet{schwarm2005reading} and \citet{petersen2006assessing}. They used statistical language models, average sentence length, the average number of syllables per word, parse features, and some other features to train and test their classification model. They also used some traditional readability scores, such as the Flesch-Kincaid readability score as an input for their model. Other studies, such as the work of \citet{kate2010learning}, used syntactical features of text to assess its readability, which augmented the accuracy of the assessment. Cohesive features were used in other research studies undertaken by \citet{sung2015constructing} and \citet{vajjala2014readability}. However, a recent study has demonstrated that a more straightforward feature like word frequency can be more significant than cohesive features on improving the text readability assessment results \cite{todirascu2016cohesive}.

Unlike traditional approaches, machine learning approaches can be applied for short texts. Models designed by \citet{vajjala2014readability} and \citet{vstajner2017automatic} are capable of assessing the readability of texts as short as a single sentence.

Another machine learning approach to text readability assessment is the ranking-based approach. In this approach, instead of classifying the text into some readability classes, a classifier is trained to compare two texts and then decide which text is more readable than the other. Having used this classifier as a comparison function, the rank-based approach sorts out all texts in a text collection according to their readability, which can be more useful for some applications. Some examples of recent research studies introducing ranking-based machine learning models for text readability assessment are the works of \citet{tanaka2010sorting}, \citet{ma2012ranking}, and \citet{vajjala2014assessing}. This research paper intends to introduce a machine learning approach toward assessing the text readability for Persian text.

\section{Dataset}
\label{Dataset}

As mentioned in section \ref{Related Works}, no research studies were available for Persian text readability assessment. Therefore, a Persian dataset for text readability assessment was collected in order to use machine learning to automate the Persian text readability assessment. The dataset was collected as a multi-class dataset to be applicable for classification models, given the better performance of classification models compared to regression models \cite{feng2010comparison}.

There are two significant approaches to text readability dataset collection. The first approach is to use texts labeled by text readability experts; the second approach is to crowdsource the information required for a text readability dataset. This research study selects the crowdsource approach since it is more accessible and can reflect the real readability of Persian texts as the labels have been determined by a vast number of Persian speakers. These labels are the Persian voters\textquotesingle\ opinion about the readability of the questioned Persian text.

The texts for the dataset were gathered from various sources and belonged to different topics. Some of the texts were gathered from Persian websites such as fa.wikipedia, beytoote.com, koodakan.org, tebyan.net, akhlagh.porsemani.ir, dastanak.com, shahrekhabar.ir, and zoomit.ir. Some texts were gathered from several Telegram\footnote{Telegram.org} messenger channels such as Sedanet, Vivaphilosophy, and Filmosophy. Finally, this research study selected some texts from several Persian books (e.g., Akhlagh Naseri by Nasir al-Din al-Tusi, Tarikh-e Jahangosha-ye Joveini by Ata Malik Joveini, Kelileh o Demneh by Ibn al-Muqaffa, and Gulistan by Saadi Shirazi). There are two selection criteria for these sources: Firstly, most of the selected sources are common sources of online information reached by Persian users. Secondly, the sources cover a wide range of general text readability levels, from children's stories to some difficult to comprehend novels. These texts cover various genres, such as news, children's stories, novels, sports, history, science, philosophy, and so forth. The number of texts in each genre is presented in Table \ref{topics-table}.

\begin{table}[htbp]
\caption{Number of texts in each topic.}
\begin{center}
\label{topics-table}       
\begin{tabular}{|c|c|}
\hline
\textbf{Topic}                                    & \textbf{Number of texts}           \\
\hline
Children stories                  & 1747                    \\
Teen stories                        & 2358                      \\
Wikipedia articles                       & 1233                      \\
News                       & 880                      \\
Tech articles                        & 780                      \\
Political articles                        & 1143                      \\
Philosophy                        & 1091                      \\
Movie reviews                        & 1708                      \\
Novels                        & 1840                      \\
\hline
\end{tabular}
\end{center}
\end{table}

In order to gather readability information from Persian speakers, the Telegram messenger platform was selected. Telegram is an open-source, cross-platform messenger that is popular among Iranians. In addition to a vast number of Telegram users in Iran, Telegram messenger is capable of hosting third-party chatbots. In order to gather information for the text readability dataset, a Telegram chatbot was designed. This chatbot asked Telegram users about Persian texts\textquotesingle\ readability and requested users to express their opinions about the readability of those texts. The user could submit his/her opinion by choosing among three options, including easy, medium, and hard readability. Three levels of readability were chosen since a higher number of classes might have confused the user to select the appropriate readability level. The reason is that it is not possible to set a clear and understandable definition for each readability level. Fewer readability levels are not suitable because the readability information collected from users would not be adequate to develop a machine learning model for text readability assessment, which can have useful applications in the real world.

In order to make sure each chatbot user has a clear understanding of each readability level, the number of unfamiliar words, grammatical complexity, text and sentence length, and overall understandability were introduced to them, and each readability level was roughly described using these criteria. Chatbot users were asked to evaluate each text based on these criteria. After designing and implementing such a chatbot, the bot was published in some popular Telegram channels so that Persian users could interact with the chatbot. Chatbot users were composed of two groups. The first group was initial collaborators, who were undergraduate college students. The second group was public Telegram users of different ages, genders, and levels of education.

The main shortcoming of collecting labels using crowdsourcing methods is the likelihood of human errors, malicious users, and most importantly, the disagreement between voters. In order to avoid such errors, three solutions were implemented. Firstly, the chatbot was designed to collect at least three labels per each text from distinct users. As shown in Table \ref{dataset-table}, an average of 3.5 labels were collected per each text in the dataset. Still, three labels can not avoid all the errors that could arise. Secondly, around one hundred gold standard texts were chosen. These gold standard texts were evaluated by all the volunteers. Utilizing these texts, it was possible to evaluate the reading ability of each volunteer, in addition to the detection of possible malicious users. The information gathered using these gold standard texts were noisy due to the limited number of gold standard texts that were possible to ask each voter as the engagement time of the user with the chatbot was brief. Each user was evaluated using nine gold standard texts, with three texts from each readability level. Thirdly, the user reading level metric was included. The user reading level is defined as the percentage of easy, medium, and hard labels provided by a voter. Practicing user reading level, it is possible to find malicious users by finding outliers in user reading levels, in addition to having a firm understanding of each voter's reading skill. This metric is further discussed in section \ref{Proposed Approach}. A vital step to ensure an accurate and quality dataset was to solely select the texts with more than 80\% agreement on their labels among voters. Since the agreement percentage was rounded down, and the average number of labels per text is 3.5, most selected texts have 100\% agreement on their labels. The selected portion of the dataset is used to test the Persian text readability assessment model represented in section \ref{Proposed Approach}. The labels were gathered in approximately three months. Some information regarding the collected dataset is presented in Table \ref{dataset-table}.

\begin{table}[!htbp]
\begin{center}
\caption{Statistics of the gathered persian text readability dataset.}
\label{dataset-table}       
\begin{tabular}{|c|c|}
\hline
\textbf{Property}                                    & \textbf{Value}           \\
\hline
Total number of texts in the dataset                 & 12780                    \\
Total number of collaborators                        & 400                      \\
Average number of texts labeled by each collaborator & 127                      \\
Average number of labels per text                    & 3.5                      \\
Average text length (word)                      & 37  \\
Average text length (characters)                 & 173 \\
Total number of labels gathered                      & 45368                    \\
Portion of easy labels                               & 54 percent               \\
Portion of medium labels                             & 32 percent               \\
Portion of hard labels                               & 14 percent              \\
\hline
\end{tabular}
\end{center}
\end{table}

\section{Proposed Approach}
\label{Proposed Approach}

One of the crucial steps of designing and implementing a machine learning model is to find and select suitable features. The present research aims to design a machine learning model to automate the text readability assessment for the Persian language. The Persian language belongs to the Indo-European language family. Note that Persian is also called Farsi. One of the distinctive properties of Farsi is the extensive use of prefixes and postfixes. This property means that different meanings can be derived by adding a prefix or postfix to a word. The Persian language also has very loose grammatical rules. It suggests expressing the same meaning by multiple and different order of words. However, all these features do not make the Persian language much different from other languages in the Indo-European language family. In order to create a useful machine learning model for the Persian text readability assessment, the features should be carefully selected. Due to the similarities of the Persian language and other Indo-European languages, it is possible to use the most beneficial features from other research studies aimed at text readability assessment for other similar languages in order to get desired results in this research. Therefore, a list of features was assembled from other related studies. Though there are more complex features proposed for similar models, studies have shown that more complex features have little contribution to the accuracy of the text readability model \cite{todirascu2016cohesive, franccois2012ai}. Consequently, the more proven features such as frequency and POS language models \cite{todirascu2016cohesive}, word and sentence length, which are a part of Flesch-Kincaid formula, and other similar features were selected to guarantee desired results and other more experimental features were scheduled for further studies. The list of selected features is reported in Table \ref{feature-table}.

\begin{table}[!htbp]
\begin{center}
\caption{The list of used features.}
\label{feature-table}       
\begin{tabular}{|c|}
\hline
\textbf{Feature}                                      \\
\hline
Average length of sentences in the text                                     \\
Variance of sentences length in the text                                    \\
Average length of sentences in the text                                     \\
Variance of words length in the text                                        \\
Average word n-gram model frequency (n = 1 to 5)                            \\
Average character n-gram model frequency (n = 1 to 5)                       \\
Variance of word n-gram model frequency (n = 1 to 5)                        \\
Variance of character n-gram model frequency (n = 1 to 5)                   \\
Number of sentences in the text                                             \\
Number of words in the text                                                 \\
Number of characters in the text                                            \\
Number of unique words in the text                                          \\
Entropy (number of unique words divided by total number of words)           \\
Average of n-max unigram model frequency words (n = 1 to 5)                 \\
Average of n-min unigram model frequency words (n = 1 to 5)                 \\
Percentage of each POS tagged words to the total number of words \\
Average n-gram POS model frequency (n = 1 to 5)                  \\
Variance of n-gram POS model frequency (n = 1 to 5)              \\
User reading ability                                                                          \\
\hline
\end{tabular}
\end{center}
\end{table}

In Table \ref{feature-table}, word N-gram is a sequence of words with the length of N, and character N-gram is a sequence of characters with the length of N.

Another statistical model used here was N-gram, which was a part-of-speech model. In order to design such a model, firstly, a version of the Hamshahri Persian corpus \cite{aleahmad2009hamshahri} was created. In this version,  every word was replaced with its part-of-speech tag. Then, a word N-gram model was created from the modified Hamshahri corpus. To calculate the average N-gram part-of-speech frequency of a text, each word was replaced with its part-of-speech tag. Then, it uses the previously created N-gram part-of-speech model to calculate the frequency of each N-gram.

All statistical language models were developed using Hamshahri Persian corpus (Table \ref{feature-table}). Also, part-of-speech tagging was executed by Hazm python library\footnote{"Sobhe/hazm: python library for digesting persian text," Aug 2017, retrieved September 1,2017. [Online]. Available: https://github.com/sobhe/hazm}.
To achieve a higher degree of accuracy, the texts underwent a set of processes such as normalization and stopword removal in the dataset.

Furthermore, the users\textquotesingle\ reading ability was taken into account in this research study in order to increase the accuracy of predicted text readability. User reading ability is defined by the portion of easy, medium, and hard texts tagged by the user from a preselected uniform set of easy, medium, and hard texts. User reading ability was extracted from the information gathered from each chatbot user. Thanks to this feature, it was possible to assess the text readability for a particular reader by identifying the reading ability of that reader. To define a user reading ability for each data point tagged by multiple chatbot users, the average reading ability of users labeling the text was used. To ensure that these reading ability levels are uniform, there were some pre-labeled texts in the chatbot which were asked from every chatbot user in order to determine his/her reading ability.

Because of the differences in features’\textquotesingle\ scales, it was essential to perform feature scaling on data points in order to enhance the accuracy of the model. This task was performed by the tools available in the Scikit-learn machine learning python library \cite{scikit-learn}. The processed features and the difficulty levels, derived from the chatbot, were then fed to classifiers such as support vector machine, linear support vector machine, random forest, decision tree, and Gaussian naive Bayes (hereafter GNB), which are available in Scikit-learn machine learning library. The test results are discussed in section \ref{Experiments}.

\section{Experiments}
\label{Experiments}

In order to test the created model, a ten-fold cross-validation technique was used. As ten-fold cross-validation indicates, in each experiment, 90 percent of labeled texts from the dataset is used for training, and the other 10 percent is used for testing. The final results were demonstrated based on precision, recall, and f1-score measures.

In the conducted experiments, support vector machine, linear support vector machine, decision tree, and Gaussian naive Bayes classifiers were used with default settings. Random forest classifier was used with 50 estimators, with the test results shown in Table \ref{overallresults-table}. The reported precision, recall, and f1-score are weighted measures, indicating that the total precision, recall, and f1-scores are a weighted average of each class\textquotesingle s precision, recall, and f1-score. The weights are the number of data points in each class. Because of the unbalanced number of data points in each class in the gathered dataset (Table \ref{dataset-table}), the effect of the precision, recall, and f1-score of each class on final results is different.

\begin{table}[!htbp]
\begin{center}
\caption{\label{overallresults-table} Classification test results using multiple classifiers. }
\begin{tabular}{|c|c|c|c|c|}
\hline
\multirow{2}{*}{\textbf{Classifier}} & \textbf{Precision} & \textbf{Recall} & \textbf{F1-score} & \textbf{ROC\textendash AUC} \\
& \textbf{(train/test)} & \textbf{(train/test)} & \textbf{(train/test)} & \textbf{(train/test)}
\\
\hline
\multirow{2}{*}{SVM} & 0.92 & 0.93 & 0.92 & 0.87 \\ \cline{2-5} 
& 0.89 & 0.89 & 0.89 & 0.83 \\
\hline
\multirow{2}{*}{} Linear & 0.9 & 0.91 & 0.9 & 0.85 \\ \cline{2-5}
SVM & 0.9 & 0.9 & 0.9 & 0.84 \\
\hline
\multirow{2}{*}{} Random & 1.0 & 1.0 & 1.0 & 1.0 \\ \cline{2-5}
forest & 0.88 & 0.89 & 0.88 & 0.81\\
\hline
\multirow{2}{*}{} Decision & 1.0 & 1.0 & 1.0 & 1.0 \\ \cline{2-5}
tree & 0.83 & 0.83 & 0.83 & 0.80 \\
\hline
\multirow{2}{*}{GBN} & 0.83 & 0.66 & 0.71 & 0.77 \\ \cline{2-5}
& 0.82 & 0.63 & 0.68 & 0.76 \\ 
\hline
\end{tabular}
\end{center}
\end{table}

As shown in Table \ref{overallresults-table}, most classifiers had high precision, recall, and f1-score. Linear support vector machine outperformed other classifiers, which yielded an f1-score of 0.9. These results suggest that this model could accurately label Persian texts by their readability level. The precision, recall, and f1-score of random forest and decision tree models in training were 1, which indicates overfitting in these models. In order to have a more in-depth insight into the performance of the classifiers, a class level classification report of SVM classifier results has been displayed in Table \ref{details-table}.

\begin{table}[!htbp]
\begin{center}
\caption{Class level test results of SVM classifier. }
\label{details-table} 
\begin{tabular}{|c|c|c|c|}
\hline
\multirow{2}{*}{\textbf{Class}}          & \textbf{Precision} & \textbf{Recall} & \textbf{F1-score} \\
& \textbf{(train/test)} & \textbf{(train/test)} & \textbf{(train/test)} \\
\hline
\multirow{2}{*}{Easy}   & 0.93                            & 0.99                         & 0.96                           \\ \cline{2-4}
                        & 0.92                            & 0.98                         & 0.95                           \\
\hline
\multirow{2}{*}{Medium} & 0.89                            & 0.59                         & 0.71                           \\ \cline{2-4}
                        & 0.85                            & 0.5                          & 0.63                           \\
\hline
\multirow{2}{*}{Hard}   & 0.92                            & 0.9                          & 0.91                           \\ \cline{2-4}
                        & 0.84                            & 0.84                         & 0.84                           \\
\hline
\end{tabular}
\end{center}
\end{table}

As reported in Table \ref{details-table}, the support vector machine model could effectively classify texts in easy and hard classes. However, the result of the medium class was different. Model\textquotesingle s precision in medium class was high, but its recall was lower than the recall of other classes.

\begin{figure}[!htbp]
\begin{subfigure}{.23\textwidth}
\centering
\includegraphics[width=\linewidth]{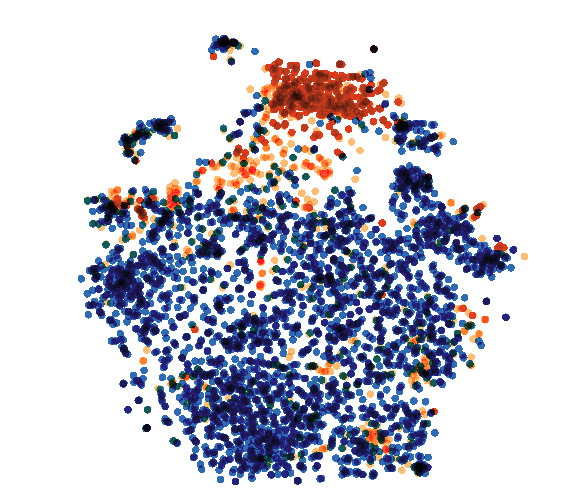}
\caption{2D visualization of all data points in the text readability dataset.}
\label{fig:all}
\end{subfigure}
\hfill
\begin{subfigure}{0.23\textwidth}
\centering
\includegraphics[width=\linewidth]{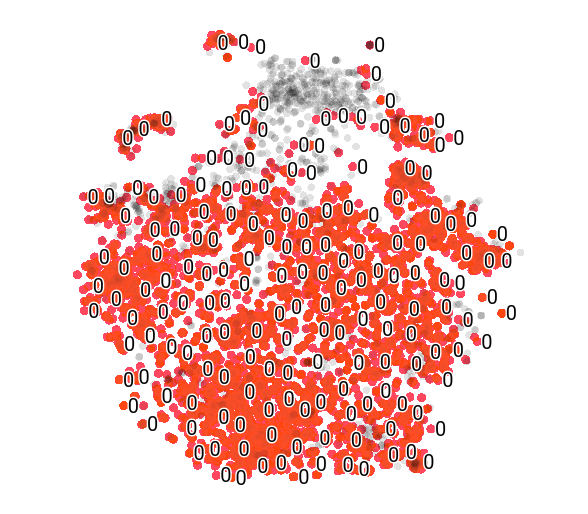}
\caption{2D visualization of the easy labeled data points in the text readability dataset.}
\label{fig:easy}
\end{subfigure}
\medskip
\begin{subfigure}{0.23\textwidth}
\centering
\includegraphics[width=\linewidth]{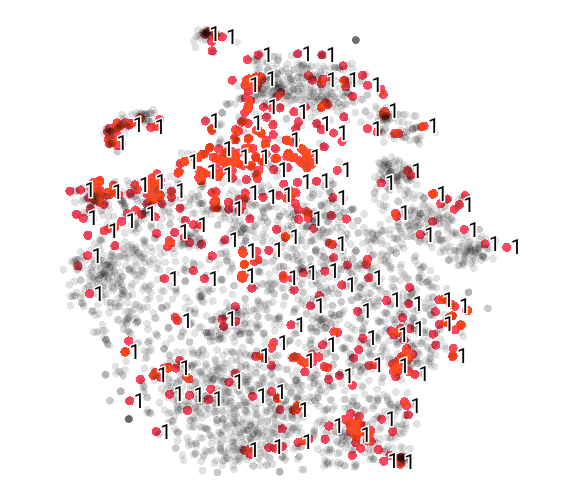}
\caption{2D visualization of the medium labeled data points in the text readability dataset.}
\label{fig:medium}
\end{subfigure}
\hfill
\begin{subfigure}{0.23\textwidth}
\centering
\includegraphics[width=\linewidth]{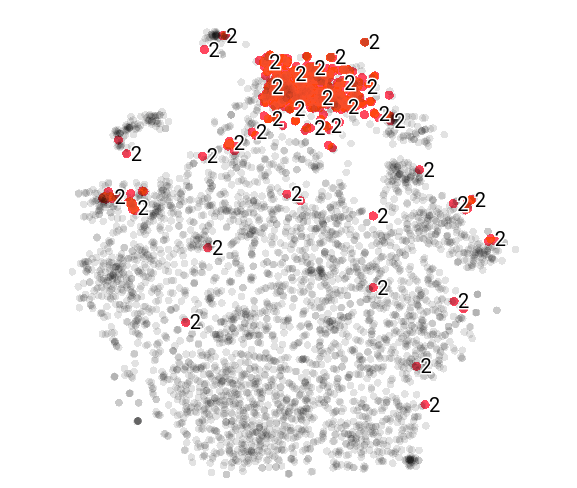}
\caption{2D visualization of the hard labeled data points in the text readability dataset.}
\label{fig:hard}
\end{subfigure}
\caption{2D visualization of the text readability dataset.}
\end{figure}

To further analyze the problem of medium class recall, the features extracted from the dataset were visualized. A visualization tool from the Tensorflow library \cite{tensorflow2015-whitepaper}, called Embedding Projector, was used to visualize the text readability dataset. The Embedding Projector employed the t-SNE \cite{maaten2008visualizing} technique to reduce the dimensionality of the dataset. The t-SNE was applied to visualize the dataset with a high number of dimensions in a 2- or 3-dimensional space. The visualization results are demonstrated in Figures \ref{fig:all} to \ref{fig:hard}.





Figure \ref{fig:all}, blue reveals easy class data points, green represents medium class data points, and red reflects hard class data points. Further, in Figures \ref{fig:easy} to \ref{fig:hard}, 0 denotes easy class data points, 1 shows the medium class data points, and 2 reveals hard class data points. As depicted in Figures \ref{fig:all} to \ref{fig:hard}, some data points from texts in the medium class were mixed with the data points of other classes. The problem of low recall in the medium class was not resolved here. Nonetheless, other studies have shown that the medium class has been highly opinion-based and been heavily dependent on a reader\textquotesingle s definition of a text with medium readability. In order to improve medium class recall, a more concrete definition or new features for the medium readability class are required to capture each reader\textquotesingle s opinion on the definition of medium class.

\section{Conclusions and Future works}
\label{Conclusions and Future works}

In this paper, two important goals were fulfilled: (i) the first Persian text readability dataset was gathered using a novel solution; and (ii) the first machine learning model for Persian text readability assessment was introduced. The machine learning model introduced in this research had high accuracy and could be employed in many applications such as text simplification, automated medical and educational text assessment, finding suitable content for second language learners, and so forth. Future research will focus on investigating and introducing new features such as LSA and TFIDF, in order to improve the accuracy of the proposed model. On top of that, the large number of texts in the gathered dataset makes it suitable for the implementation of deep learning models on text readability assessment, which could be another interesting future study.

\bibliographystyle{IEEEtranN}
\bibliography{mybibs}

\begin{thebibliography}{32}
\providecommand{\natexlab}[1]{#1}
\providecommand{\url}[1]{#1}
\csname url@samestyle\endcsname
\providecommand{\newblock}{\relax}
\providecommand{\bibinfo}[2]{#2}
\providecommand{\BIBentrySTDinterwordspacing}{\spaceskip=0pt\relax}
\providecommand{\BIBentryALTinterwordstretchfactor}{4}
\providecommand{\BIBentryALTinterwordspacing}{\spaceskip=\fontdimen2\font plus
\BIBentryALTinterwordstretchfactor\fontdimen3\font minus
  \fontdimen4\font\relax}
\providecommand{\BIBforeignlanguage}[2]{{%
\expandafter\ifx\csname l@#1\endcsname\relax
\typeout{** WARNING: IEEEtranN.bst: No hyphenation pattern has been}%
\typeout{** loaded for the language `#1'. Using the pattern for}%
\typeout{** the default language instead.}%
\else
\language=\csname l@#1\endcsname
\fi
#2}}
\providecommand{\BIBdecl}{\relax}
\BIBdecl

\bibitem[Xia et~al.(2019)Xia, Kochmar, and Briscoe]{xia2019text}
M.~Xia, E.~Kochmar, and T.~Briscoe, ``Text readability assessment for second
  language learners,'' \emph{arXiv preprint arXiv:1906.07580}, 2019.

\bibitem[Leroy et~al.(2013)Leroy, Kauchak, and Mouradi]{leroy2013user}
G.~Leroy, D.~Kauchak, and O.~Mouradi, ``A user-study measuring the effects of
  lexical simplification and coherence enhancement on perceived and actual text
  difficulty,'' \emph{International Journal of Medical Informatics}, vol.~82,
  no.~8, pp. 717--730, aug 2013.

\bibitem[Pancer et~al.(2019)Pancer, Chandler, Poole, and
  Noseworthy]{pancer2019readability}
E.~Pancer, V.~Chandler, M.~Poole, and T.~J. Noseworthy, ``How readability
  shapes social media engagement,'' \emph{Journal of Consumer Psychology},
  vol.~29, no.~2, pp. 262--270, 2019.

\bibitem[Kincaid et~al.(1975)Kincaid, Fishburne, P., L., and
  S.]{kincaid1975derivation}
J.~P. Kincaid, J.~Fishburne, R.~R. P., C.~R. L., and B.~S., ``Derivation of new
  readability formulas (automated readability index, fog count and flesch
  reading ease formula) for navy enlisted personnel,'' Tech. Rep., feb 1975.

\bibitem[Heydari and Riazi(2012)]{heydari2012readability}
P.~Heydari and A.~M. Riazi, ``Readability of texts: Human evaluation versus
  computer index,'' \emph{Mediterranean Journal of Social Sciences}, vol.~3,
  no.~1, pp. 177--190, 2012.

\bibitem[Begeny and Greene(2013)]{begeny2014can}
J.~C. Begeny and D.~J. Greene, ``{CAN} {READABILITY} {FORMULAS} {BE} {USED}
  {TO} {SUCCESSFULLY} {GAUGE} {DIFFICULTY} {OF} {READING} {MATERIALS}?''
  \emph{Psychology in the Schools}, vol.~51, no.~2, pp. 198--215, nov 2013.

\bibitem[Crossley et~al.(2017)Crossley, Skalicky, Dascalu, McNamara, and
  Kyle]{crossley2017predicting}
S.~A. Crossley, S.~Skalicky, M.~Dascalu, D.~S. McNamara, and K.~Kyle,
  ``Predicting text comprehension, processing, and familiarity in adult
  readers: New approaches to readability formulas,'' \emph{Discourse
  Processes}, vol.~54, no. 5-6, pp. 340--359, mar 2017.

\bibitem[Schwarm and Ostendorf(2005)]{schwarm2005reading}
S.~E. Schwarm and M.~Ostendorf, ``Reading level assessment using support vector
  machines and statistical language models,'' in \emph{Proceedings of the 43rd
  Annual Meeting on Association for Computational Linguistics}.\hskip 1em plus
  0.5em minus 0.4em\relax Association for Computational Linguistics, 2005, pp.
  523--530.

\bibitem[Dayani(2000)]{dayani2000criteria}
M.~Dayani, ``A criteria for assessing the persian texts’ readability,''
  \emph{Journal of Social Science and Humanities}, vol.~10, pp. 35--48, 2000.

\bibitem[Flesch(1943)]{flesch1943marks}
R.~Flesch, ``Marks of readable style; a study in adult education.''
  \emph{Teachers College Contributions to Education}, 1943.

\bibitem[Stenner(1996)]{stenner1996measuring}
A.~J. Stenner, ``Measuring reading comprehension with the lexile framework.''
  1996.

\bibitem[Collins-Thompson and Callan(2005)]{collins2005predicting}
K.~Collins-Thompson and J.~Callan, ``Predicting reading difficulty with
  statistical language models,'' \emph{Journal of the American Society for
  Information Science and Technology}, vol.~56, no.~13, pp. 1448--1462, 2005.

\bibitem[Petersen and Ostendorf(2009)]{petersen2009machine}
S.~E. Petersen and M.~Ostendorf, ``A machine learning approach to reading level
  assessment,'' \emph{Computer Speech {\&} Language}, vol.~23, no.~1, pp.
  89--106, jan 2009.

\bibitem[Hartley(2016)]{hartley2016time}
J.~Hartley, ``Is time up for the flesch measure of reading ease?''
  \emph{Scientometrics}, vol. 107, no.~3, pp. 1523--1526, mar 2016.

\bibitem[Kidwell et~al.(2011)Kidwell, Lebanon, and
  Collins-Thompson]{kidwell2011statistical}
P.~Kidwell, G.~Lebanon, and K.~Collins-Thompson, ``Statistical estimation of
  word acquisition with application to readability prediction,'' \emph{Journal
  of the American Statistical Association}, vol. 106, no. 493, pp. 21--30, mar
  2011.

\bibitem[Petersen and Ostendorf(2006)]{petersen2006assessing}
S.~E. Petersen and M.~Ostendorf, ``Assessing the reading level of web pages,''
  in \emph{Ninth International Conference on Spoken Language Processing}, 2006.

\bibitem[Feng et~al.(2009)Feng, Elhadad, and Huenerfauth]{feng2009cognitively}
L.~Feng, N.~Elhadad, and M.~Huenerfauth, ``Cognitively motivated features for
  readability assessment,'' in \emph{Proceedings of the 12th Conference of the
  European Chapter of the Association for Computational Linguistics on - {EACL}
  {\textquotesingle}09}.\hskip 1em plus 0.5em minus 0.4em\relax Association for
  Computational Linguistics, 2009.

\bibitem[Fran{\c{c}}ois and Miltsakaki(2012)]{franccois2012nlp}
T.~Fran{\c{c}}ois and E.~Miltsakaki, ``Do nlp and machine learning improve
  traditional readability formulas?'' in \emph{Proceedings of the First
  Workshop on Predicting and Improving Text Readability for target reader
  populations}.\hskip 1em plus 0.5em minus 0.4em\relax Association for
  Computational Linguistics, 2012, pp. 49--57.

\bibitem[Feng et~al.(2010)Feng, Jansche, Huenerfauth, and
  Elhadad]{feng2010comparison}
L.~Feng, M.~Jansche, M.~Huenerfauth, and N.~Elhadad, ``A comparison of features
  for automatic readability assessment,'' in \emph{Coling 2010: Posters}.\hskip
  1em plus 0.5em minus 0.4em\relax Coling 2010 Organizing Committee, 2010, pp.
  276--284.

\bibitem[Kate et~al.(2010)Kate, Luo, Patwardhan, Franz, Florian, Mooney,
  Roukos, and Welty]{kate2010learning}
R.~Kate, X.~Luo, S.~Patwardhan, M.~Franz, R.~Florian, R.~Mooney, S.~Roukos, and
  C.~Welty, ``Learning to predict readability using diverse linguistic
  features,'' in \emph{Proceedings of the 23rd International Conference on
  Computational Linguistics (Coling 2010)}.\hskip 1em plus 0.5em minus
  0.4em\relax Coling 2010 Organizing Committee, 2010, pp. 546--554.

\bibitem[Sung et~al.(2014)Sung, Chen, Cha, Tseng, Chang, and
  Chang]{sung2015constructing}
Y.-T. Sung, J.-L. Chen, J.-H. Cha, H.-C. Tseng, T.-H. Chang, and K.-E. Chang,
  ``Constructing and validating readability models: the method of integrating
  multilevel linguistic features with machine learning,'' \emph{Behavior
  Research Methods}, vol.~47, no.~2, pp. 340--354, apr 2014.

\bibitem[Vajjala and Meurers(2015)]{vajjala2014readability}
S.~Vajjala and D.~Meurers, ``Readability assessment for text simplification:
  From analysing documents to identifying sentential simplifications,''
  \emph{Recent Advances in Automatic Readability Assessment and Text
  Simplification}, vol. 165, no.~2, pp. 194--222, jan 2015.

\bibitem[Todirascu et~al.(2016)Todirascu, Fran{\c{c}}ois, Bernhard, Gala, and
  Ligozat]{todirascu2016cohesive}
A.~Todirascu, T.~Fran{\c{c}}ois, D.~Bernhard, N.~Gala, and A.-L. Ligozat, ``Are
  cohesive features relevant for text readability evaluation?'' in \emph{26th
  International Conference on Computational Linguistics (COLING 2016)}, 2016,
  pp. 987--997.

\bibitem[Stajner et~al.(2017)Stajner, Ponzetto, and
  Stuckenschmidt]{vstajner2017automatic}
S.~Stajner, S.~P. Ponzetto, and H.~Stuckenschmidt, ``Automatic assessment of
  absolute sentence complexity,'' in \emph{Proceedings of the Twenty-Sixth
  International Joint Conference on Artificial Intelligence}.\hskip 1em plus
  0.5em minus 0.4em\relax International Joint Conferences on Artificial
  Intelligence Organization, aug 2017.

\bibitem[Tanaka-Ishii et~al.(2010)Tanaka-Ishii, Tezuka, and
  Terada]{tanaka2010sorting}
K.~Tanaka-Ishii, S.~Tezuka, and H.~Terada, ``Sorting texts by readability,''
  \emph{Computational Linguistics}, vol.~36, no.~2, pp. 203--227, jun 2010.

\bibitem[Ma et~al.(2012)Ma, Fosler-Lussier, and Lofthus]{ma2012ranking}
Y.~Ma, E.~Fosler-Lussier, and R.~Lofthus, ``Ranking-based readability
  assessment for early primary children's literature,'' in \emph{Proceedings of
  the 2012 Conference of the North American Chapter of the Association for
  Computational Linguistics: Human Language Technologies}.\hskip 1em plus 0.5em
  minus 0.4em\relax Association for Computational Linguistics, 2012, pp.
  548--552.

\bibitem[Vajjala and Meurers(2014)]{vajjala2014assessing}
S.~Vajjala and D.~Meurers, ``Assessing the relative reading level of sentence
  pairs for text simplification,'' in \emph{Proceedings of the 14th Conference
  of the European Chapter of the Association for Computational
  Linguistics}.\hskip 1em plus 0.5em minus 0.4em\relax Association for
  Computational Linguistics, 2014.

\bibitem[Fran{\c{c}}ois and Fairon(2012)]{franccois2012ai}
T.~Fran{\c{c}}ois and C.~Fairon, ``An ai readability formula for french as a
  foreign language,'' in \emph{Proceedings of the 2012 Joint Conference on
  Empirical Methods in Natural Language Processing and Computational Natural
  Language Learning}.\hskip 1em plus 0.5em minus 0.4em\relax Association for
  Computational Linguistics, 2012, pp. 466--477.

\bibitem[AleAhmad et~al.(2009)AleAhmad, Amiri, Darrudi, Rahgozar, and
  Oroumchian]{aleahmad2009hamshahri}
A.~AleAhmad, H.~Amiri, E.~Darrudi, M.~Rahgozar, and F.~Oroumchian, ``Hamshahri:
  A standard persian text collection,'' \emph{Knowledge-Based Systems},
  vol.~22, no.~5, pp. 382--387, jul 2009.

\bibitem[Pedregosa et~al.(2011)Pedregosa, Varoquaux, Gramfort, Michel, Thirion,
  Grisel, Blondel, Prettenhofer, Weiss, Dubourg, Vanderplas, Passos,
  Cournapeau, Brucher, Perrot, and Duchesnay]{scikit-learn}
F.~Pedregosa, G.~Varoquaux, A.~Gramfort, V.~Michel, B.~Thirion, O.~Grisel,
  M.~Blondel, P.~Prettenhofer, R.~Weiss, V.~Dubourg, J.~Vanderplas, A.~Passos,
  D.~Cournapeau, M.~Brucher, M.~Perrot, and E.~Duchesnay, ``Scikit-learn:
  Machine learning in {P}ython,'' \emph{Journal of Machine Learning Research},
  vol.~12, pp. 2825--2830, 2011.

\bibitem[Abadi et~al.(2015)Abadi, Agarwal, Barham, Brevdo, Chen, Citro,
  Corrado, Davis, Dean, Devin, Ghemawat, Goodfellow, Harp, Irving, Isard, Jia,
  Jozefowicz, Kaiser, Kudlur, Levenberg, Man\'{e}, Monga, Moore, Murray, Olah,
  Schuster, Shlens, Steiner, Sutskever, Talwar, Tucker, Vanhoucke, Vasudevan,
  Vi\'{e}gas, Vinyals, Warden, Wattenberg, Wicke, Yu, and
  Zheng]{tensorflow2015-whitepaper}
\BIBentryALTinterwordspacing
M.~Abadi, A.~Agarwal, P.~Barham, E.~Brevdo, Z.~Chen, C.~Citro, G.~S. Corrado,
  A.~Davis, J.~Dean, M.~Devin, S.~Ghemawat, I.~Goodfellow, A.~Harp, G.~Irving,
  M.~Isard, Y.~Jia, R.~Jozefowicz, L.~Kaiser, M.~Kudlur, J.~Levenberg,
  D.~Man\'{e}, R.~Monga, S.~Moore, D.~Murray, C.~Olah, M.~Schuster, J.~Shlens,
  B.~Steiner, I.~Sutskever, K.~Talwar, P.~Tucker, V.~Vanhoucke, V.~Vasudevan,
  F.~Vi\'{e}gas, O.~Vinyals, P.~Warden, M.~Wattenberg, M.~Wicke, Y.~Yu, and
  X.~Zheng, ``{TensorFlow}: Large-scale machine learning on heterogeneous
  systems,'' 2015, software available from tensorflow.org. [Online]. Available:
  \url{https://www.tensorflow.org/}
\BIBentrySTDinterwordspacing

\bibitem[Maaten and Hinton(2008)]{maaten2008visualizing}
L.~v.~d. Maaten and G.~Hinton, ``Visualizing data using t-sne,'' \emph{Journal
  of machine learning research}, vol.~9, no. Nov, pp. 2579--2605, 2008.

\end{thebibliography}

\end{document}